\documentclass{article}

\PassOptionsToPackage{numbers, compress}{natbib}
\usepackage[final]{nips_2017}

\usepackage[utf8]{inputenc} 
\usepackage[T1]{fontenc}    
\usepackage{hyperref}       
\usepackage{url}            
\usepackage{booktabs}       
\usepackage{amsfonts}       
\usepackage{nicefrac}       
\usepackage{microtype}      
\usepackage{amsmath,amsfonts,amsthm,bm} 
\usepackage{algorithm}
\usepackage[noend]{algpseudocode}
\usepackage[textsize=scriptsize,textwidth=3cm]{todonotes}
\usepackage{caption}
\usepackage{subcaption}
\usepackage{pgfplots}
\newcommand{\thetahat}{\ensuremath{\hat{\theta}}}
\newcommand{\thetatilde}{\ensuremath{\tilde{\theta}}}
\newcommand{\R}{\ensuremath{\mathbb{R}}}

\title{Graph-Sparse Logistic Regression}

\author{
  Alexander LeNail\\
  MIT Biological Engineering\\
  \texttt{lenail@mit.edu} \\
  \And
  Ludwig Schmidt\\
  MIT CSAIL\\
  \texttt{ludwigs@mit.edu} \\
  \AND
  Johnathan Li\\
  MIT Biological Engineering\\
  \texttt{iamjli@mit.edu} \\
  \And
  Tobias Ehrenberger \\
  MIT Biological Engineering\\
  \texttt{tobieh@mit.edu} \\
  \And
  Karen Sachs \\
  MIT Biological Engineering\\
  \texttt{karens@alum.mit.edu} \\
  \And
  Stefanie Jegelka \\
  MIT CSAIL\\
  \texttt{stefje@mit.edu} \\
  \And
  Ernest Fraenkel \\
  MIT Biological Engineering\\
  \texttt{fraenkel-admin@mit.edu} \\
}

\begin{document}

\maketitle

\begin{abstract}
We introduce Graph-Sparse Logistic Regression, a new algorithm for classification for the case in which the support should be sparse but connected on a graph. We validate this algorithm against synthetic data and benchmark it against L1-regularized Logistic Regression. We then explore our technique in the bioinformatics context of proteomics data on the interactome graph. We make all our experimental code public and provide GSLR as an open source package.\protect\footnote{\href{https://github.com/fraenkel-lab/gslr}{https://github.com/fraenkel-lab/GSLR}}
\end{abstract}

\section{Introduction}

In many scientific applications of machine learning, the goal is not only to build an accurate predictor, but also to learn more about the underlying process modeled by the predictor. In such cases, the machine learning model must serve the dual objectives of accuracy and interpretability. Perhaps the simplest way of achieving interpretability is to constrain the model to select a small subset of the features to predict the label. This is interpretability by model scale: a small model is easier to interpret than a large one. 

In the case of linear classifiers, the canonical technique is L1-regularization, i.e. "the Lasso"\citep{lasso} which adds a penalty term to the objective function proportional to the L1-norm of the classifier weights. 

In some cases, this first-approximation solution will not satisfactorily select the relevant features, despite selecting features which form a support for a high-accuracy predictor. Many scientific domains are rife with subtle, complex data derived from subtle, complex processes by noisy, biased instruments, and in some of those cases the Lasso is underspecified for the task. Consider for example the common bioinformatics feature selection problem of pathway prediction. In this setting, we measure gene expression for each gene in the genome, or protein abundances for each protein in the proteome, for patients drawn from disease and control populations, and the goal is to determine which genes/proteins are pertinent to the disease. An influential study showed that some 90\% of random sets of 100 features were predictive of the label in a Breast Cancer dataset, demonstrating the difficulty of selecting the correct feature set exclusively from the information in the labels \citep{mostRandomGeneSetsPreidctBRCA}.

However, in some domains, we have side information about the features, which we can leverage for the selection of a more meaningful support. For the pathway inference problem, we have information about these proteins' relationships, which we can represent as a graph. These Protein-Protein Interaction (PPI) networks (alternately, "interactomes") draw an edge between proteins if they physically interact, weighted by the confidence in the interaction \citep{InBioMap}. We anticipate that biologically relevant features will be tightly connected in the interactome. This presents the need for an algorithm which predicts class labels from a small support which is connected on the feature graph. 

\section{Related Work}

\subsection{Sparse Regressions}

There exist many algorithmic variants of the Lasso designed to take into account structure in the feature-space, notably the Group Lasso \citep{GroupLasso} and its variants \citep{SparseGroupLasso,GroupLassoWithOverlaps}. However, these methods require a predefined partitioning or grouping of the variables. Two approaches extend Group Lasso to graphs: "Graph Lasso"\citep{GroupLassoWithOverlaps} proposes using graph motifs, e.g. every k-clique or k-line-graph as the grouping of features for input to Group Lasso with Overlaps, but this is somewhat inflexible (in that the grouping is still predetermined, the regression step cannot permute the groups) and is not necessarily tractable for larger graphs for all choices of groupings. Sparse Regression Incorporating Graphical Structure Among Predictors (SRIG) \citep{SRIG} is the special case of Group Lasso with Overlaps (and Graph Lasso) where the groups are the 1-hop neighborhoods of every node in the graph. Another idea, Graph-Laplacian-Regularized Logistic Regression \citep{logitLapnet} combines an L1 penalty on each of the nodes with an L2 penalty on the graph spectrum. 

\subsection{Prize-Collecting Steiner Forest for Pathway Prediction}
\label{sec:pcsf}

An approach to pathway prediction comes from combinatorial optimization. In the Prize-Collecting Steiner Forest (PCSF) formulation \citep{oi1,murat}, we seek a subgraph of the interactome which maximizes vertex prize and minimizes edge cost by minimizing the following objective:

\begin{equation}
	f(F) = \beta \sum_{v\notin V_F} p(v) + \sum_{e\in E_F} c(e) + \omega \cdot \kappa, 
\end{equation}

where $F(V_F, E_F)$ is the selected subgraph, $p(v)$ is the prize of vertex $v$, $c(e)$ is the cost of edge $e$. The reader will note that in the case where the prizes are set to some uniform, arbitrarily large value, this evaluates to the more common $MST$ problem. In the biological context, we set the cost of an edge $c(e)$ as the inverse confidence in the interaction between those proteins, and the vertex prize $p(v)$ as the average differential expression of the associated protein between case and control, scaled by hyperparameter $\beta$. PCSF has been used to generate experimentally-validated pathways in a number of disease contexts including recently medulloblastoma \citep{medullo_oi} and neurodegenerative diseases \citep{neuro_oi}. 

\subsection{Graph Sparsity}

Graph sparsity (as utilized in this paper) was first proposed as a general framework for structured sparsity in \citep{HZM11}.
In the context of regression problems, graph sparsity can be defined as follows.
There is a known graph $G = (V, E)$ defined on the features, i.e., each feature corresponds to a node $v \in V$.
We then say that a vector $\theta \in \R^d$ is $(G,s)$-\emph{graph-sparse} if the support $S$ of $\theta$ (i.e., the indices with non-zero coefficients) satisfies the following two properties: (i) The cardinality of the support is at most $s$.
(ii) The nodes corresponding to elements in $S$ form a connected subgraph of $G$, i.e., for every pair of indices $i, j \in S$ there is a path from $i$ to $j$ so that all intermediate indices are also contained in $S$ and all pairs of consecutive nodes are connected by an edge in $E$.

While \cite{HZM11} also introduced a polynomial-time algorithm for some regimes of graph-sparse regression, their method did not easily scale to large datasets.
To address this issue, \cite{ludwig} provided faster algorithms for projections onto the set of graph-sparse vectors that run in time which is \emph{nearly-linear} in the size of the graph.
Our work leverages the fast projection algorithms introduced in \citep{ludwig}.

Both \cite{HZM11} and \cite{ludwig} focus on the regular linear regression setting and the goal is to estimate an unknown vector in $\ell_2$-distance.
In contrast, we focus on logistic regression and aim to identify the most important subgraphs in the feature graph.

\section{Graph-Sparse Logistic Regression}

The motivation for Graph-Sparse Logistic Regression (GSLR) is to utilize prior knowledge of graph structure defined on the features (in our case, the interactome) in order to improve over classical (sparse) logistic regression. The majority of the signal separating positive from negative examples in our datasets is dispersed over the interactome and hence less biologically relevant. This algorithm intends to optimally separate our data with the constraint that the signals used for classification must be nearby in graph-space. Viewed otherwise, the objective is to find the subgraph of the interactome most predictive of the phenotype of interest, rather than a disjunct set of features.

\begin{algorithm}[t]
\caption{Graph-Sparse Logistic Regression}
\label{alg:gslr}
\begin{algorithmic}[1]
\Function{GSLR}{$X$, $y$, $G$, $s$, $\eta$, $k$}
\State Let $f(X, y, \theta)$ be the logistic loss.
\State $\thetahat^0 \gets 0$
\For{$i \gets 0, \ldots, k-1$}
\State $\thetatilde^{i+1} \gets \thetahat^i - \eta \cdot \nabla f(X, y, \thetahat^i)$
\State $\thetahat^{i+1} \gets P_{G,s}(\thetatilde^{i+1})$ \Comment{Graph-sparse projection}
\EndFor
\State \textbf{return} $\thetahat^k$
\EndFunction
\end{algorithmic}
\end{algorithm}

At a high level, GSLR is an instantiation of projected gradient descent with the (multiclass) logistic loss function.
In each iteration of the procedure (see Algorithm \ref{alg:gslr}),
we take a gradient step and then project the current iterate back to the constraint set.
Since projecting onto the set of graph-sparse vectors \emph{exactly} is NP-hard\footnote{This follows from hardness results for the classical Steiner tree problem.}, we
instead resort to the \emph{approximate} projections from \cite{ludwig}.
In particular, we assume we have access to a projection operator $P_{G,s}$ with the following guarantee.

Given an arbitrary vector $p \in \R^d$, the projection operator $P_{G,s}$ returns a vector $q \in \R^d$ satisfying the following two properties:
\begin{itemize}
\item \textbf{Approximate projection:} The vector $q$ is an approximate projection, i.e., instead of achieving the smallest distance to the input point $p$ among points in the constraint set, the distance achieved by $q$ is within a small constant factor:
\begin{equation}
\label{eq:approxproj}
  \|p - q \|_2^2  \; \leq \; 2 \cdot \min_{q' \textnormal{ is } (G,s)\textnormal{-sparse}} \| p - q'\|_2^2 \; .
\end{equation}
\item \textbf{Approximate graph sparsity:} The support of the vector $q$ forms a connected component of size at most $6s + 1$ in the graph $G$.
\end{itemize}

Interestingly, the projection operator $P_{G,s}$ relies on algorithmic tools originally developed for the PCSF problem (see Section \ref{sec:pcsf}).
We project onto the set of graph-sparse vectors by constructing a carefully tuned sequence of PCSF instances in which the edge costs in these instances are adjusted so that the resulting PCSF solution has the desired sparsity.
The node prizes are given by the squared coefficient corresponding to each index, so that the sum of node prizes not in the solution subtree becomes the approximation error term $\| p - q \|_2^2$ in Equation \eqref{eq:approxproj} above.
We refer the reader to \citep{ludwig} for a more detailed explanation of the approximate projection for graph sparsity.

We remark the approximation guarantees mentioned above are proven in a worst case setting.
In our experiments, we typically find that we can adjust the sparsity to within about 10\%.

\section{Experiments}

\subsection{Synthetic Data Generation}

In biomedical data, ground truth can be hard to come by. In complex diseases, the features relevant to a disease are usually not fully known, and are occasionally disputed. 

In order to circumvent the absence of ground-truth, we construct synthetic datasets by the following procedure: We model the 206 samples from \cite{OV} as a multivariate gaussian over the protein values, from which we draw new samples (synthetic "patients"). We then translate some of those samples by a "perturbation vector" and designate the translated  subset as the positive set. In order to make synthetic data which matches our biological intuition, the perturbation vector is small relative to the variance of of the gaussian. 

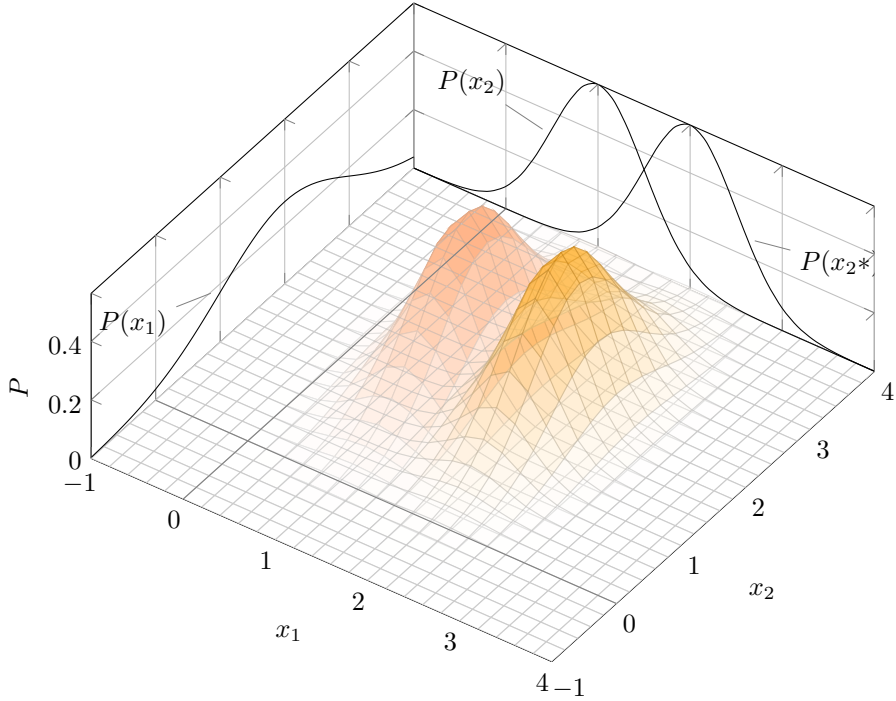
\begin{figure}[h]
  \centering
\begin{tikzpicture}[
declare function={mu1=1;},
declare function={mu2=2;},
declare function={mu3=2;},
declare function={sigma1=0.5;},
declare function={sigma2=1;},
declare function={normal(\m,\s)=1/(2*\s*sqrt(pi))*exp(-(x-\m)^2/(2*\s^2));},
declare function={bivar(\ma,\sa,\mb,\sb)=
3.3/(2*pi*\sa*\sb) * exp(-((x-\ma)^2/\sa^2 + (y-\mb)^2/\sb^2))/2;}]
\begin{axis}[
colormap name=whitered,
width=12cm,
view={35}{65},
enlargelimits=false,
grid=major,
domain=-1:4,
y domain=-1:4,
samples=26,
xlabel=$x_1$,
ylabel=$x_2$,
zlabel={$P$},
]

\addplot3 [opacity=1,surf,colormap={whitered}{color(0cm)=(white); color(1cm)=(orange!75!red)}] {bivar(mu1,sigma1,mu2,sigma2)};
\addplot3 [opacity=0.5,surf,colormap={whiteyellow}{color(0cm)=(white); color(1cm)=(orange!75!yellow)}] {bivar(mu3,sigma1,mu2,sigma2)};

\addplot3 [domain=-1:4,samples=31, samples y=0, thin, smooth] (x,4,{normal(mu1,sigma1)});
\addplot3 [domain=-1:4,samples=31, samples y=0, thin, smooth] (-1,x,{normal(mu2,sigma2)});
\addplot3 [domain=-1:4,samples=31, samples y=0, thin, smooth] (x,4,{normal(mu3,sigma1)});

\draw [black!50] (axis cs:-1,0,0) -- (axis cs:4,0,0);
\draw [black!50] (axis cs:0,-1,0) -- (axis cs:0,4,0);

\node at (axis cs:-1,1,0.18) [pin=195:$P(x_1)$] {};
\node at (axis cs:0.5,4,0.32) [pin=140:$P(x_2)$] {};
\node at (axis cs:2.6,4,0.26) [pin=-3:$P(x_2*)$] {};
\end{axis}
\end{tikzpicture}
\caption{A low dimensional cartoon of our synthetic data generation strategy. Here, the pink gaussian represents our original data, from which we sample our negative examples. We sample our positive examples from the orange gaussian by first sampling them from the pink gaussian and translating them by the perturbation vector (in this case $<0, {\mu_x}_2 - {\mu_x}_2 *>$, with $x_2$ in the pathway and $x_1$ not.}
\end{figure}

We generate perturbation vectors in two schemes. In both schemes, we pick a KEGG \cite{KEGG} pathway (a canonical, well studied group of proteins which interact in order to accomplish some biological function) and sample an offset for each of the proteins in the pathway. The perturbation vector $\vec{x}$ is only non-zero for the pathway proteins, and hence, we are "perturbing" a biological function. This provides us with ground truth, since we know the identities of the nodes we are translating.

$negative = \mathcal{N} (\vec{\mu}, \bm{\Sigma} ) \quad positive = \mathcal{N} (\vec{\mu}, \bm{\Sigma} ) + \vec{x}$

In the first scheme, we sample each pathway protein's offset from the univariate gaussian defined by that protein's empirical values in the data. 

$positive_{scheme 1} = \mathcal{N} (\vec{\mu}, \bm{\Sigma} ) + \vec{x}$ where $\vec{x_p} = \mathcal{N} (0, \sigma_p^2 )$ if $p \in$ KEGG pathway, 0 otherwise.

In expectation the perturbation vector will be 0, and so few of the pathway proteins will appear perturbed. This perturbation strategy only significantly alters a small fraction of the pathway means, which matches our biological intuition that only a few proteins in a pathway need be severely dysregulated for the pathway to be dysregulated.

In the second scheme, we sample each pathway protein's offset from a univariate gaussian centered one standard deviation from 0, causing a larger magnitude perturbation vector than in the first scheme. 

$positive_{scheme 2} = \mathcal{N} (\vec{\mu}, \bm{\Sigma} ) + \vec{x}$ where $\vec{x_p} = \mathcal{N} (\pm\sigma_p, \sigma_p^2 ) $ if protein $p \in$ KEGG pathway, 0 otherwise.

In both cases, the shift in the multivariate gaussian which defines the "positive" set is fairly subtle, imperceptible to dimensionality reduction techniques such at PCA or t-SNE.  We draw 100 positive and 100 negative samples for each of 229 pathways with an average of 80 proteins "perturbed" of 16,349 in our interactome graph.

\subsection{Synthetic Data Results}

We first examine the performance of classic L1-regularized logistic regression \citep{sklearn}. We perform 10-fold cross validation on every dataset for each of 16 sparsity hyperparameter values in order to recover sets of nonzero coefficients of varying sparsity. We can then evaluate these 10 x 229 x 16 runs in their classification accuracy, precision, and recall, where precision is the fraction of nonzero weights which correspond to true pathway proteins and recall is the fraction of true pathway proteins for which the algorithm assigns nonzero weights. Note that these are precision and recall defined for feature selection, which are different than the conventional definitions. 

\begin{figure}[h]
\centering
\begin{subfigure}{.5\textwidth}
  \centering
  \includegraphics[width=\linewidth]{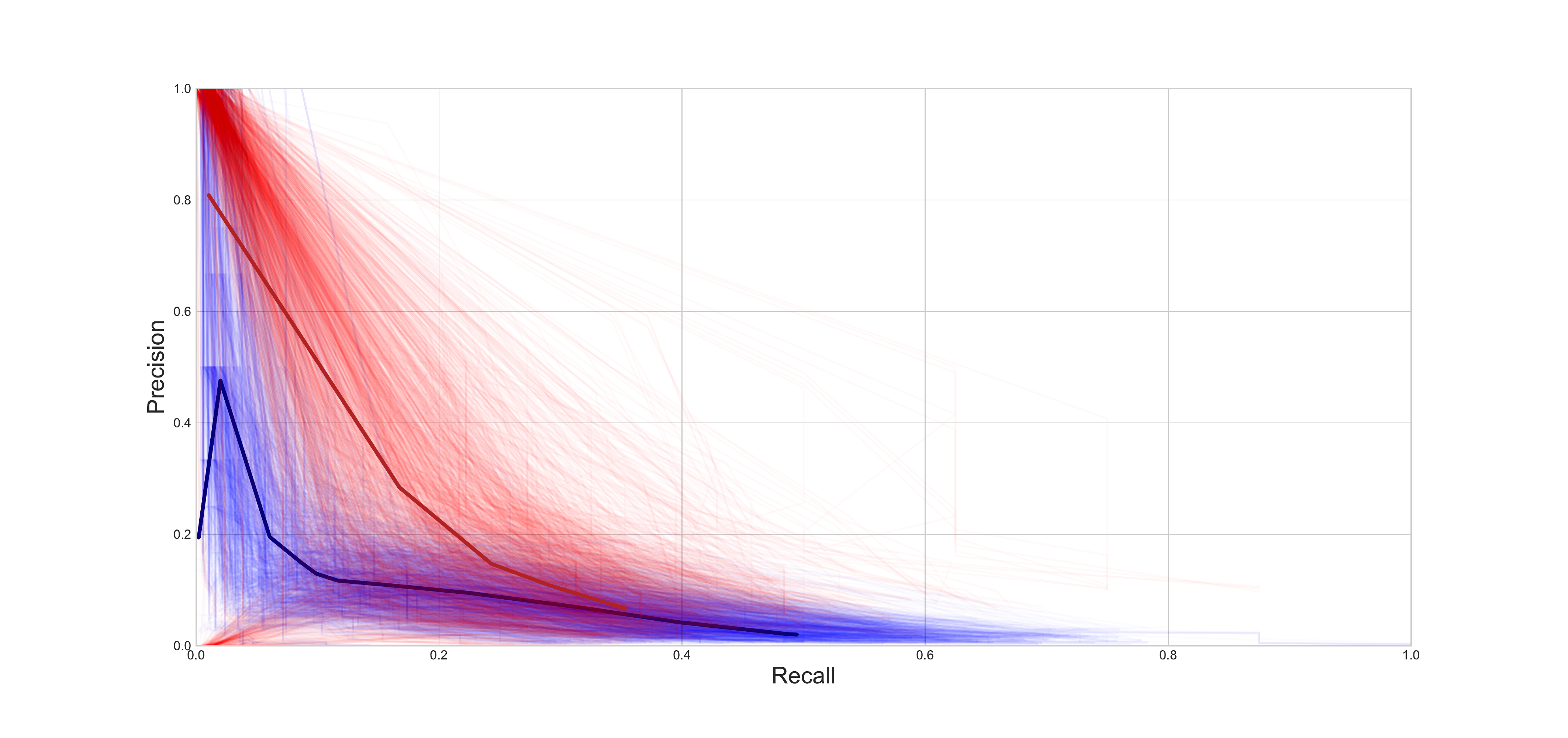}
  \caption{scheme 1 -- all datasets, all folds}
  \label{fig:sub1}
\end{subfigure}%
\begin{subfigure}{.5\textwidth}
  \centering
  \includegraphics[width=\linewidth]{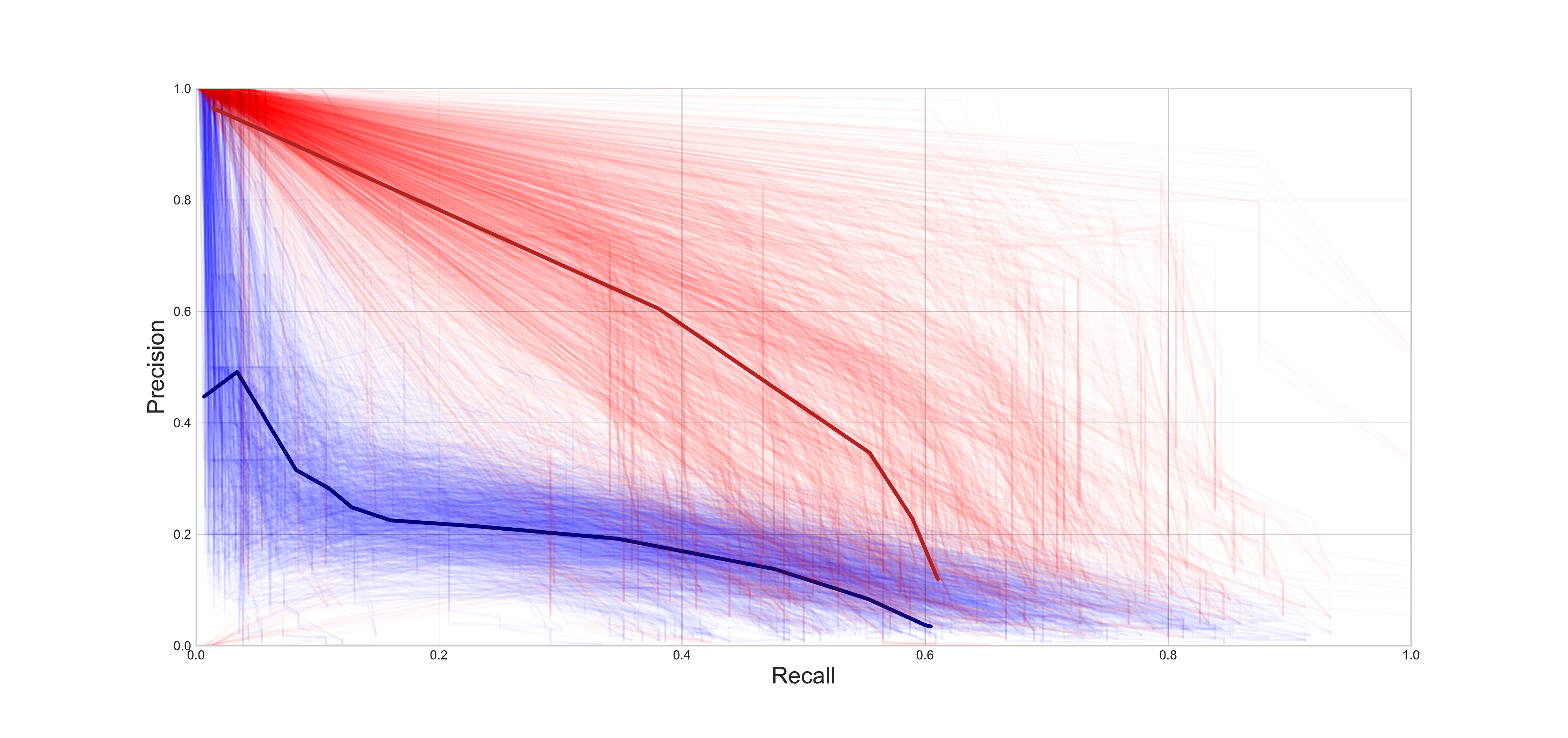}
  \caption{scheme 2 -- all datasets, all folds}
  \label{fig:sub2}
\end{subfigure}
\begin{subfigure}{.5\textwidth}
  \centering
  \includegraphics[width=\linewidth]{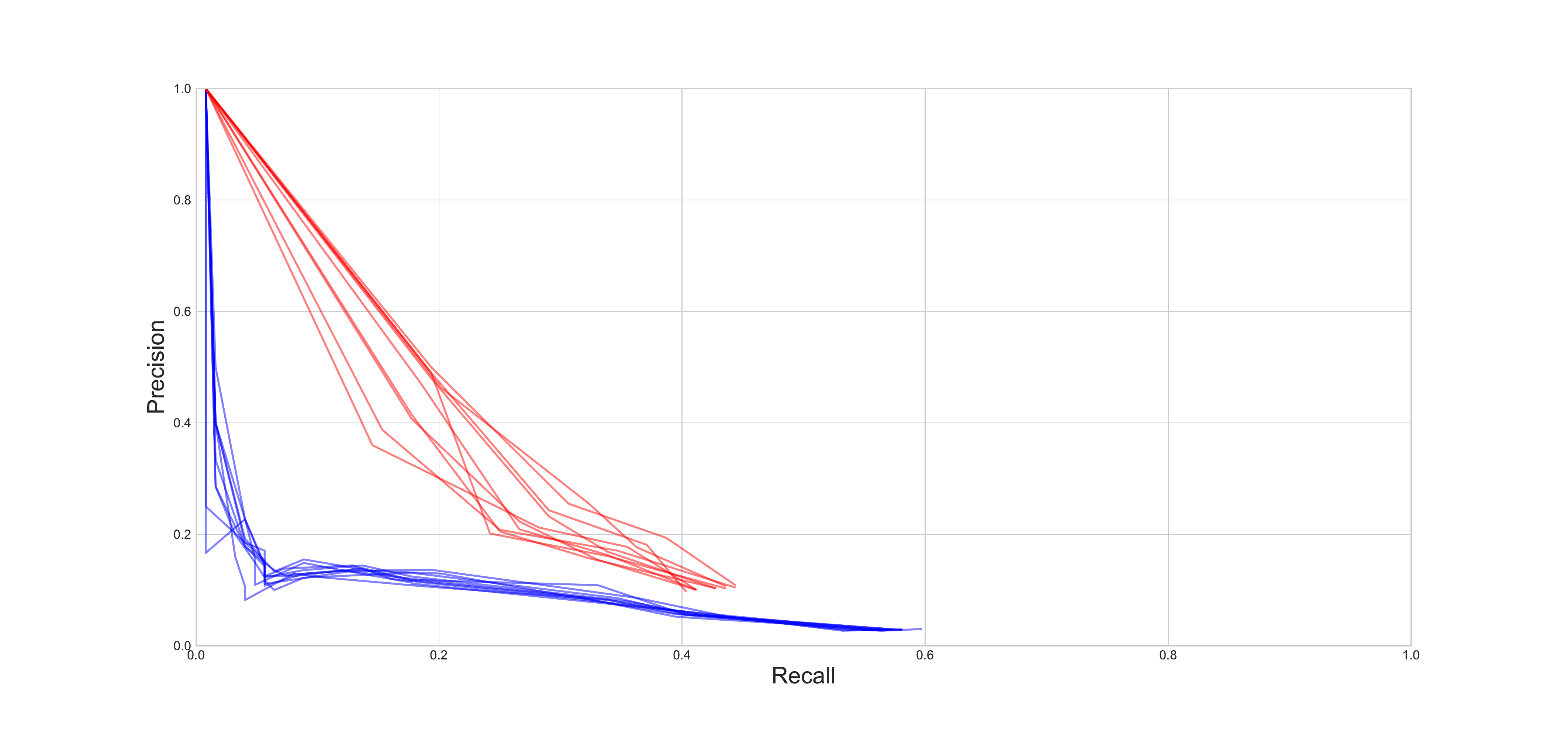}
  \caption{scheme 1 -- a random dataset, all folds}
  \label{fig:sub3}
\end{subfigure}%
\begin{subfigure}{.5\textwidth}
  \centering
  \includegraphics[width=\linewidth]{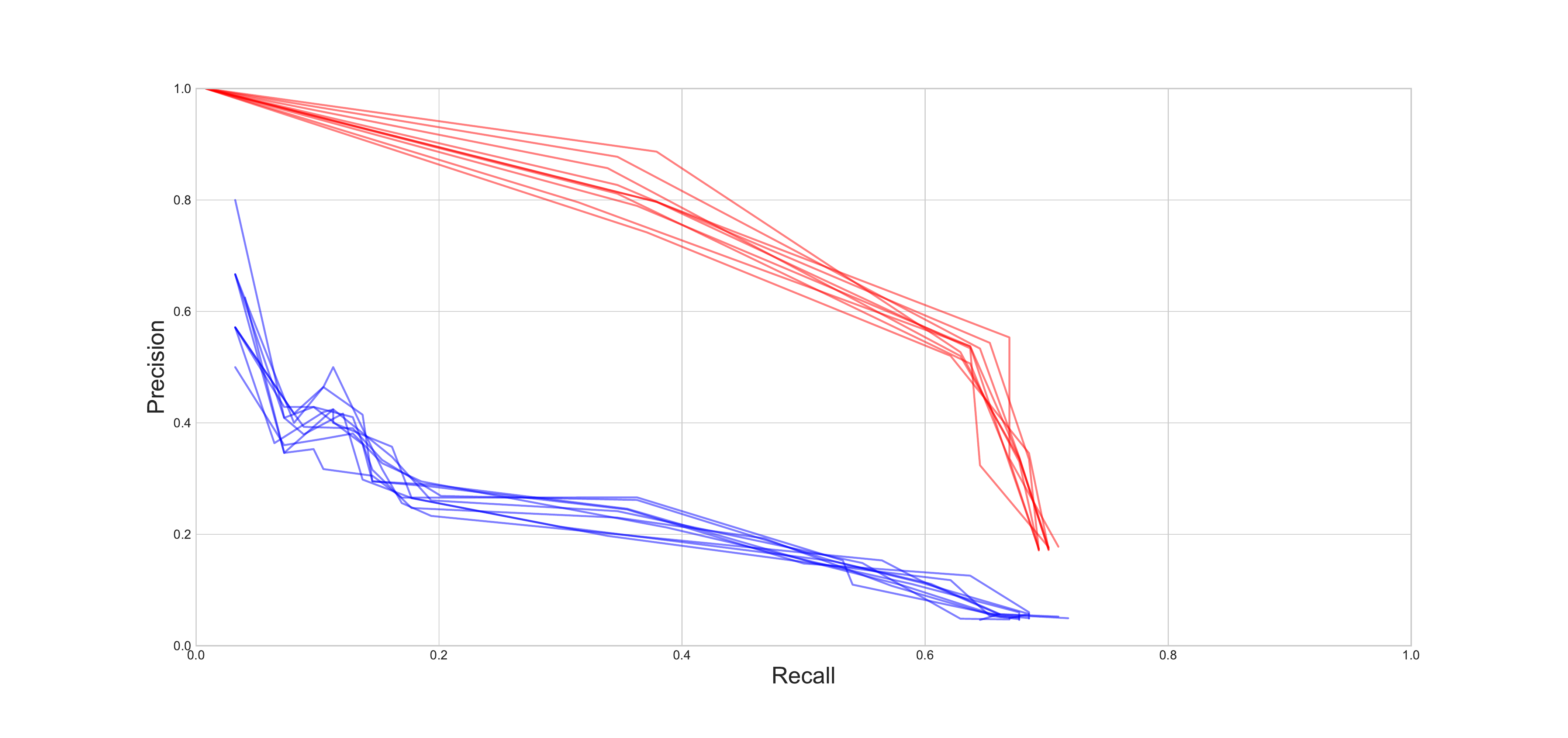}
  \caption{scheme 2 -- a random dataset, all folds}
  \label{fig:sub4}
\end{subfigure}
\caption{L1-Regularized Logistic Regression (blue) versus Graph-Sparse logistic Regression (red). Each trace represents one fold of one dataset, varying sparsity. The bolded trace is the average.}
\label{fig:test}
\end{figure}

Logistic Regression with strong L1 regularization achieves high classification accuracies on the holdout sets across pathways, regularizations, and folds, but low precision and recall (Fig 1), which is to say the set of features Logistic Regression uses to separate positive and negative examples are mostly non-overlapping with the set of true pathway proteins whose means were translated at dataset construction. In the case of ($n << d$) it is to be anticipated that true signal is near-impossible to capture without the aid of some heuristic or side-information.

We then examine the performance of Graph-Sparse Logistic Regression across 10 folds on every dataset for each of 5 sparsities, and find that it outperforms regular sparse Logistic Regression on precision and recall across the board. GSLR has nearly equivalent predictive accuracy of sparse logistic regression, but is able to more precisely select more of the true perturbed features.

\subsection{Real Data Results}

We then test our technique on the real biological data we had used to generate our synthetic data. The CPTAC ovarian cancer study \cite{OV} focuses primarily on high-grade serous ovarian carcinomas (HGSOC) which account for the majority of ovarian cancers diagnosed and are associated with the lowest survival rates in ovarian cancer. Here, we are using the findings associated with the five proteomic subtypes descibed in \citep{OV}: differentiated, metabolic, proliferative, mesenchymal, and stromal. 

We first determine which known biological processes are well represented in the GSLR-selected set of proteins using hypergeometric overlap tests with gene sets from the Molecular Signature Database (MSigDb). In concordance with \cite{OV}, we find a strong association between the stromal subtype and the complement system and (other) blood-cell related gene sets, as well as the metabolic subtype and cytokine signaling. Similarly, gene sets associated with extracellular matrix proteins ranked highly for both the mesenchymal subtype and the stromal group. The primary signal that was associated with the proliferative subtype, DNA replication, was not directly linked to this subtype by our method. We suspect that this could be because DNA replication is a signal that is shared across the subtypes and was therefore not pulled out as a subtype-distinguishing feature. Finally, for the differentiated subtype, we identified gene sets closely related to cell-cell interactions. 
We then examine the overlap between the GSLR-selected proteins and an independent list of all genes with supporting evidence of an association with Ovarian Cancer, and find that GSLR's prediction has a precision of 0.52 and a recall of 0.09. A hypergeometric test of significance gives us a p-value of 1e-68.

Given the close correspondence of the previously published subtype-associated pathways and processes and the gene sets identified by GSLR, we conclude that GSLR is suited to examine real-world biological datasets and able to assist in generating potentially clinically relevant results.

\section{Discussion}

We have introduced a logistic regression model for settings where the support of the unknown parameter vector is graph-sparse.
When features have a known graphical structure, this approach leverages this structure to select features that are both predictive and connected on the underlying graph structure.
We validated our method on synthetic data and then applied it to real proteomics data where we found that GSLR was able to select features associated with biological pathways previously found to be enriched in Ovarian Cancer subtypes.

An interesting avenue for future work is applying our technique to other domains with graphical structure in the feature space.
On the theoretical side, an important question is under what conditions GSLR is guaranteed to identify relevant features.
Finally, we would like to extend our approach to also incorporate non-linear structure in the data.

\small

\bibliography{bib.bib}

\end{document}